\pdfoutput=1

\documentclass[11pt]{article}

\usepackage[preprint]{acl}
\usepackage{pifont}  
\newcommand{\cmark}{\ding{51}}  
\newcommand{\xmark}{\ding{55}} 
\usepackage{amsmath} 
\usepackage{times}
\usepackage{amsmath}
\usepackage{latexsym}
\usepackage{colortbl}   
\usepackage{xcolor}     
\usepackage{multirow}   
\usepackage{booktabs}   
\usepackage{graphicx}   
\usepackage{amsmath}    
\usepackage{colortbl, xcolor} 
\usepackage{placeins} 
\usepackage{subcaption} 
\definecolor{lightblue}{rgb}{0.68, 0.85, 0.9} 
\definecolor{babyblue}{rgb}{0.54, 0.81, 0.94}
\usepackage{times}
\usepackage{latexsym}
\usepackage{colortbl}   
\usepackage{xcolor}     
\usepackage{graphicx}       
\usepackage{booktabs}       
\usepackage{multirow}       
\usepackage[T1]{fontenc}

\usepackage[utf8]{inputenc}
\usepackage{booktabs}  

\usepackage{microtype}

\usepackage{inconsolata}
\usepackage[utf8]{inputenc}  
\usepackage[T1]{fontenc}     
\usepackage{graphicx}

\usepackage{amsmath}
\usepackage{multirow}
\usepackage{booktabs}
\usepackage{enumitem}
\usepackage{amssymb}
\usepackage[T1]{fontenc}

\usepackage[utf8]{inputenc}
\usepackage{amsmath}
\usepackage{amssymb}
\usepackage[utf8]{inputenc} 
\usepackage{CJKutf8}         
\usepackage{inconsolata}
\usepackage[utf8]{inputenc}
\usepackage{graphicx}

\title{CLIPErase: Efficient Unlearning of Visual-Textual Associations in CLIP}

\author{
  Tianyu Yang$^1$, 
  Lisen Dai$^2$, 
  Xiangqi Wang$^1$, 
  Minhao Cheng$^3$, \\
  \textbf{Yapeng Tian}$^4$, 
  \textbf{Xiangliang Zhang}$^1$\thanks{Corresponding author: \texttt{xzhang33@nd.edu}} \\
  \\
  $^1$University of Notre Dame \\
  $^2$Columbia University \\
  $^3$Pennsylvania State University \\
  $^4$University of Texas at Dallas \\
}

\begin{document}
\maketitle
\begin{abstract}

Machine unlearning (MU) has gained significant attention as a means to remove the influence of specific data from a trained model without requiring full retraining. While progress has been made in unimodal domains like text and image classification, unlearning in multimodal models remains relatively under-explored. In this work, we address the unique challenges of unlearning in CLIP, a prominent multimodal model that aligns visual and textual representations. We introduce CLIPErase, a novel approach that disentangles and selectively forgets both visual and textual associations, ensuring that unlearning does not compromise model performance.
CLIPErase consists of three key modules: a Forgetting Module that disrupts the associations in the forget set, a Retention Module that preserves performance on the retain set, and a Consistency Module that maintains consistency with the original model. Extensive experiments on CIFAR-100, Flickr30K, and Conceptual 12M across five CLIP downstream tasks, as well as an evaluation on diffusion models, demonstrate that CLIPErase effectively removes designated associations from multimodal samples in downstream tasks, while preserving the model’s performance on the retain set after unlearning. The project's code is available at: \href{https://tianyu-yang-anna.github.io/ClipErase-ACL/}{\textit{https://tianyu-yang-anna.github.io/ClipErase-ACL/}}.

\end{abstract}

\section{Introduction}

\begin{figure}[!htbp]
\centering
\includegraphics[width=0.5\textwidth]{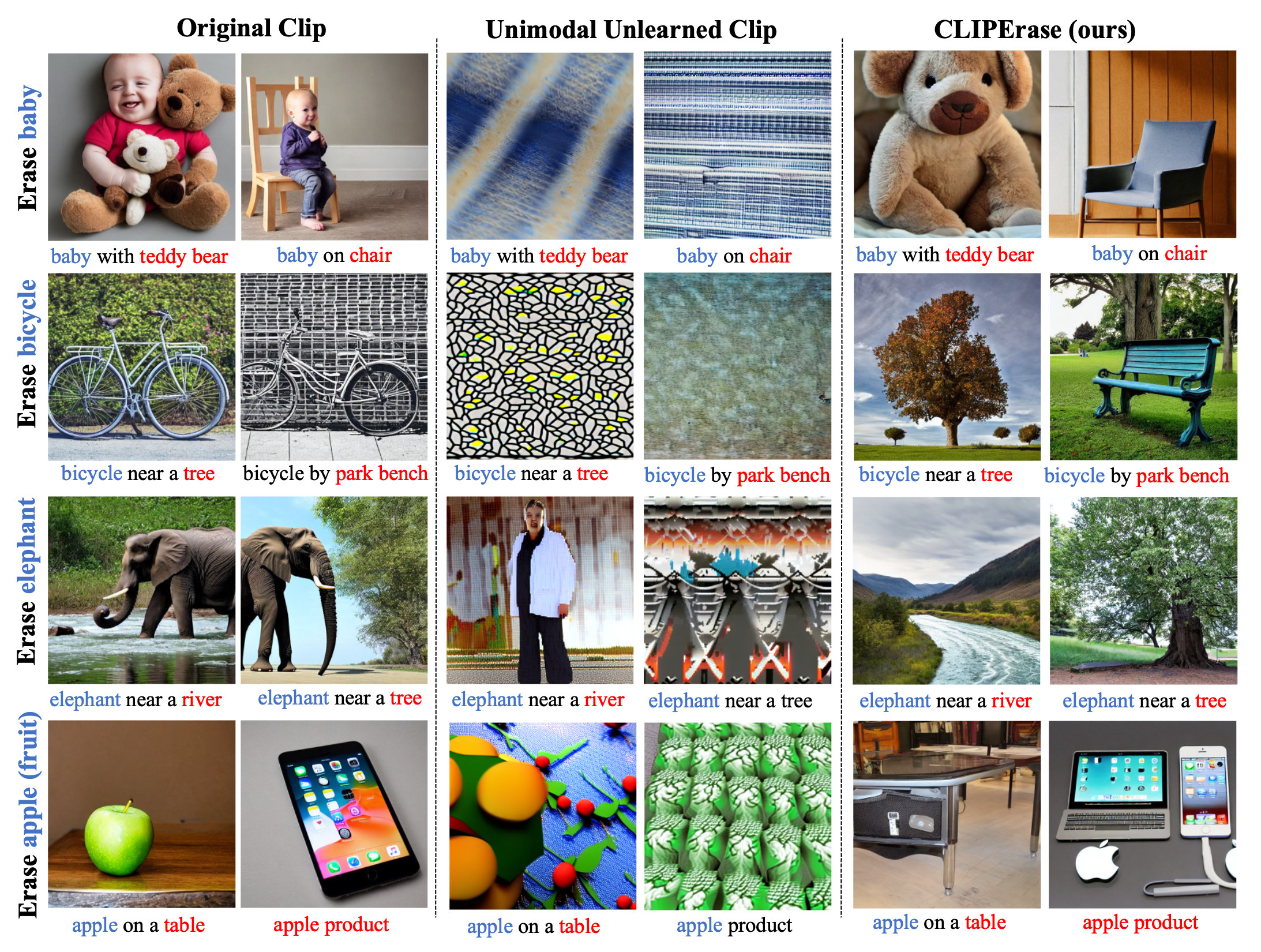}
\caption{
Comparison of Stable Diffusion results using original CLIP, unimodal unlearned CLIP (with Gradient Ascent on the text modality), and our CLIPErase shows that unimodal unlearning introduces distortions and fails to remove targeted concepts, whereas CLIPErase selectively erases them and preserves other details.
}
\vspace{-2mm}
\label{intro}
\end{figure}
\vspace{-1.0mm}

Multimodal models \cite{Kim2021DiffusionCLIPTD,liu2024visual,yuan2021florence,zhai2022lit,li2023blip,li2022blip} such as CLIP have shown powerful representational capabilities in tasks such as image-text retrieval and text-to-image generation. However, as these models continue to evolve and expand their applicability, the need for multimodal machine unlearning (MU) becomes increasingly urgent. This is because large-scale multimodal training datasets often contain sensitive or copyrighted content whose influence must be removed from the model’s learned representations,
whether to comply with new regulations, protect user privacy, or address intellectual property concerns. 


While MU has made notable progress in  removing or altering features within single-modality domains like images and text. However, its application to multimodal models remains largely unexplored.  These multimodal models learn from interconnected modalities, such as images and text, and represent them in a shared embedding space.  Consequently, unimodal MU approaches, which perturb features in only one modality, can unintentionally disrupt crucial cross-modal relationships. This disruption negatively impacts downstream tasks, particularly those relying on precisely learned text-image alignment, and can even render the resulting embeddings unusable. As shown in the first three rows of Figure~\ref{intro}, applying unimodal unlearning to CLIP can compromise embeddings for tasks like image generation with diffusion models, resulting in either a failure to generate meaningful images or an inability to remove the targeted concept.

Moreover, multimodal data often exhibits complex associations, where a single word may correspond to multiple concepts across different modalities.  For example, the word "apple" could refer to either the tech company or the fruit.  Unimodal unlearning methods lack the precision to selectively forget a dedicated concept. Given "apple" as the unlearning target, these approaches would erase all meanings associated with "apple," as shown in Row 4 of Figure~\ref{intro}, instead of targeting a specific sense, such as the fruit.

To address these challenges, we introduce CLIPErase, a novel multimodal unlearning framework specifically designed for pretrained CLIP models. Instead of indiscriminately erasing entire concepts, CLIPErase selectively removes specific associations while preserving other learned cross-modal semantic correspondences, thereby preventing disruption of the shared embedding space. This approach offers two key advantages, as demonstrated in Figure~\ref{intro}. First, CLIPErase can "forget" designated concepts without harming unrelated ones like "chairs" or "Teddy Bears". This contrasts with unimodal methods, which often remove these unrelated concepts inadvertently or even fail to generate any concepts. Second, CLIPErase can selectively remove specific associations within multiple mappings. For instance, it can remove the association of "apple" with the fruit while preserving its association with Apple products like iPhones. 

To precisely refine and control the knowledge stored within CLIP models, 
CLIPErase consists of three core components: \textbf{(1) Forgetting Module} disrupts the associations between images and text in the forget set by minimizing their cross-modal similarity, effectively removing the targeted connections; \textbf{(2) Retention Module} preserves performance on the retain set by maintaining contrastive alignment, preventing unintended damage to the shared embedding space;  \textbf{(3) Consistency Module} maintains consistency by penalizing deviations in the unimodal (text and image) distributions compared to the original model. These three modules work together to efficiently perform unlearning, eliminating the need for retraining from scratch. 
Our key contributions are highlighted as follows: \vspace{-0.15cm}\begin{itemize}[leftmargin=*,itemsep=0pt,parsep=0pt]
\item We introduce CLIPErase, an innovative framework designed for unlearning pretrained CLIP models, enabling efficient removal of specific multimodal associations without retraining.\vspace{-0.01cm} 
\item CLIPErase utilizes three modules to disrupt multimodal data alignment for targeted unlearning while preserving performance on the retain set.\vspace{-0.01cm} 
\item We demonstrate the impact and applicability  of CLIPErase through extensive experiments across five downstream tasks, including zero-shot prediction, retrieval, and integration with diffusion models for image generation.
\end{itemize}

\begin{figure*}[h!]
\centering
\includegraphics[width=1\textwidth]{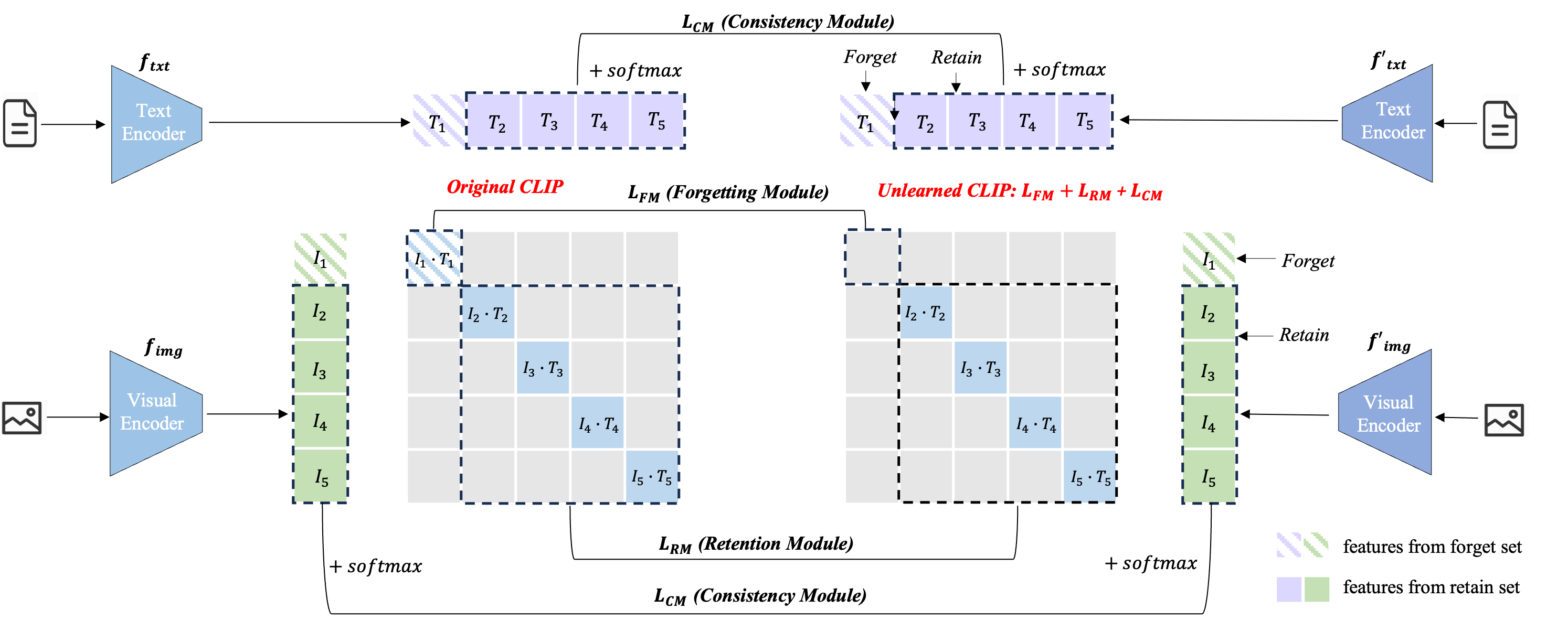}
\vspace{-1mm}
\caption{Overview of the CLIPErase framework, consisting of   three key modules: (a) Forgetting Module: disrupts cross-modal associations within the forget set to weaken the undesired image and text associations; (b) Retention Module: preserves cross-modal associations within the retain set; (c) Consistency Module: maintains consistency with the original model by aligning unimodal representations.
} 

\label{fig1}
\end{figure*}
\section{Related Works}
As machine learning models continue to grow in size and their training datasets become increasingly vast and complex, the concept of MU has garnered significant attention in academic and industry~\cite{mcconnon2024machineunlearning, pedregosa2023unlearningchallenge} to promoting AI for social welfare. MU aims to selectively remove specific information—such as private data~\cite{Zhang2023FedRecoveryDP}, outdated knowledge~\cite{Wang2023KGAAG}, and harmful content~\cite{Liu2024TowardsSL} from a trained model without necessitating a complete retraining from scratch~\cite{Bourtoule2019MachineU}. We next discuss these techniques in the context of text and image modalities.

\noindent
\textbf{Machine Unlearning in Text:} MU has been extensively studied in the text domain. \citet{gupta2021adaptive} first explored adaptive parameter tuning, while \citet{maini2024tofu}, \citet{chen2023unlearn}, and \citet{jia2024soul} leveraged gradient-based methods, including second-order optimization and KL-Divergence descent. \citet{eldan2023s} focused on preference optimization. To address high computational costs, \citet{kurmanji2024towards} proposed scalable methods to approximate the unlearning process, while \citet{chen2023boundary} aimed to directly reduce overhead. \citet{ullah2021machine} employed differential privacy to enhance trust in MU and derived theoretical guarantees.

\noindent
\textbf{Machine Unlearning in Vision:}  
Previous works on MU in vision focus on eliminating the influence of specific visual data in classification models~\cite{kurmanji2024towards} and unlearning visual patterns in diffusion models ~\cite{zhang2024forget,gandikota2024unified, gandikota2023erasing,fuchi2024erasing}. 

Recently,  a few studies emerged to study how to unlearn multimodal embedding models. 
\citet{cheng2023multidelete} introduced a multi-deletion mechanism via modality decoupling, but it relies on randomly sampled unrelated pairs, which can fail on small or imbalanced datasets. Similarly, \citet{kravets2024zero} used Lipschitz regularization and synthetic data for zero-shot forgetting, but face high computational costs and data quality issues. However, both methods are task-specific, focusing on either retrieval or zero-shot tasks. In contrast, our approach unlearns the targeted visual and textual associations within CLIP itself, allowing a wider range of downstream tasks to benefit, such as zero-shot image classification, image/text retrieval, and diffusion-based image generation.

\section{Preliminary}
\textbf{The CLIP Model}. 
CLIP is a multimodal pre-trained model that learns to align images and their text descriptions in a shared embedding space. It consists of an image encoder $f_{\text{img}}$ and a text encoder $f_{\text{txt}}$. Given a training dataset $\{(x_i^n, x_t^n)\}_{n=1}^N$, where $x_i^n$ represents the $n$-th image and $x_t^n$ its corresponding textual description, CLIP projects these inputs into a common latent space:   $f_{\text{img}}(x_i^n)\in \mathbb{R}^d$ and $f_{\text{txt}}(x_t^n)\in \mathbb{R}^d$, which are normalized feature embeddings for images and texts, respectively.

CLIP is trained using a contrastive loss that optimizes alignment between matched image-text pairs while pushing apart mismatched pairs. For a batch of \( N \) image-text pairs, the loss is defined as: 
\vspace{-0.05in}
\begin{equation}
\small
\mathcal{L} = -\frac{1}{N} \sum_{n=1}^N \log \text{softmax}  (   f_{\text{img}}(x_i^n) \cdot f_{\text{txt}}(x_t^n)/\tau  )
\label{eq:contrastive_loss} \vspace{-0.05in}
\end{equation}
where \( \tau \) is a temperature parameter, and the softmax function normalizes the similarity scores across all text samples in the batch, converting them into a probability distribution.

\noindent
\textbf{Problem Definition}.
Consider a pre-trained CLIP model \( \Theta \) trained on a dataset \( D \). 
Our goal is to design an unlearning algorithm that removes the influence of a specified  forget set  \( D_f \) from \( \Theta \) without degrading the model's performance on the remaining data, referred to as the retain set \( D_r = D - D_f \). Let  \( \Theta_u \) be the model undergoing this unlearning process.  We aim to achieve that 
in the unlearned model \( \Theta_u \), images and their corresponding text descriptions in \( D_f \) should no longer be aligned. This means that in downstream tasks like image retrieval, an image from \( D_f \) will no longer be retrieved by its original associated text description, and vice versa. Meanwhile, the unlearning process should not compromise the model's ability to effectively align images and text from the retain set \( D_r \). 

\section{Proposed Method CLIPErase}

Unlearning in multimodal models like CLIP presents unique challenges. Unlike unimodal models, where information is encoded in a single modality, CLIP relies on the intricate interplay between visual and textual features within a shared embedding space. Consequently, removing the influence of specific data necessitates more than simply adjusting one encoder in isolation.  We need a mechanism to precisely disrupt the cross-modal associations learned from the forget set \( D_f \) while leaving the associations from the retain set \( D_r \) intact. Traditional MU methods, primarily designed for unimodal settings,  are ill-equipped for this task, as they fail to account for the complex relationship between image and text representations in CLIP.

To address the unique challenges of unlearning in CLIP, we introduce CLIPErase, a novel unlearning algorithm designed specifically for pre-trained CLIP models. As illustrated in Figure~\ref{fig1}, CLIPErase modifies both the image and text encoders of the original CLIP model,  \( \Theta \), by integrating three core modules: Forgetting Module, Retention Module, and Consistency Module.

\subsection{Forgetting Module}

Pre-trained CLIP models capture rich semantic relationships between images and text, making it challenging to induce forgetting of specific concepts without disrupting other learned associations. To address this, we introduce a forget module designed to selectively weaken the image-text binding within the forget set \( D_f \). This module operates intentionally by misaligning the visual and textual representations corresponding to \( D_f \), ensuring that the model no longer recognizes the cross-modal relationships present in the discarded data.

The forget module is guided by the following optimization objective, \( \mathcal{L}_{\text{FM}} \), which minimizes the similarity between image and text features within the forget set. Specifically,   \( \mathcal{L}_{\text{FM}} \) is defined as:
\vspace{-1mm}
\begin{equation}
\small
\mathcal{L}_{\text{FM}} = \frac{1}{N_f} \sum_{n=1}^{N_f} \left( f_{\text{img}}(x_i^n) \cdot f_{\text{txt}}(x_t^n) \right)
\label{eq:fm_loss}
\end{equation}
where \( N_f \) is the number of samples in \( D_f \). 
By minimizing the raw dot product between the image and text embeddings, we directly disrupt their alignment.  This simple approach effectively drives the dot product towards zero or negative values, ensuring that images and text from \( D_f \) no longer retrieve each other in downstream tasks.

\subsection{Retention Module}

While the forget module focuses on disrupting the alignment of image-text pairs in the forget set, this process can inadvertently affect the model's performance on the retain set \( D_r \). This is because adjustments to the model's parameters can affect the overall cross-modal representation space, influencing the model's understanding and processing of the retain set.


To mitigate this, we introduce a retention module designed to preserve the model's performance on \( D_r \) during the unlearning process by minimizing the alignment loss:  
\begin{equation}
\small
\mathcal{L}_{\text{RM}} = -\frac{1}{N_r} \sum_{n=1}^{N_r} \log \text{softmax}  (   f_{\text{img}}(x_i^n) \cdot f_{\text{txt}}(x_t^n)/{\tau}  )
\label{eq:rm_loss}
\end{equation}
where \( N_r \) is the number of samples in \( D_r \). 
This is the same as the original CLIP contrastive loss function. This choice is motivated by the fact that the contrastive loss effectively maintains the desired image-text alignments within \( D_r \). Other loss functions, such as mean squared error (MSE) on the embeddings, would not adequately preserve the structured, pairwise relationships that are fundamental to CLIP's functionality. With contrastive loss, we ensure that each image in \( D_r \) remains closely aligned with its corresponding text, while being distinct from other image-text pairs. This strategy effectively preserves the model's intended behavior on the retain set.  Furthermore, by leveraging CLIP's original training objective, we avoid introducing conflicting learning signals that could hinder the retention of the desired associations.

\begin{table*}[ht]
\centering 
\renewcommand{\arraystretch}{1}  
\scalebox{0.85}{ 
\begin{tabular}{c|c|c}
\hline
\textbf{Task} & \textbf{Dataset} & \textbf{Metric} \\ \hline
Zero-shot Prediction \& Retrieval & CIFAR-100,Conceptual 12M   & Accuracy of $D_f$$\downarrow$, $D_r$$\uparrow$ \\ 
Image \& Text Retrieval & Flickr30k  & Recall (R@1, R@5, R@10) of $D_f$$\downarrow$, $D_r$$\uparrow$ \\ 
Diffusion  & Flickr30k  &  Detection Rate\\ 
\hline
\end{tabular}
}
\caption{Experimental Setup: Evaluation Tasks, Datasets, Metrics}
\label{setting1}
\end{table*}

\subsection{Consistency Module}
The distinct optimization objectives applied to the forget set \( D_f \) and the retain set \( D_r \) may introduce unexpected errors or biases in the model's predictions on \( D_r \). To mitigate this risk, we introduce a consistency module that encourages the unlearned model \( \Theta_u \) to maintain similar behavior to the original model \( \Theta \) on the retain set. This is achieved by adding a consistency regularization term, \( \mathcal{L}_{\text{CM}} \), defined as the Kullback-Leibler (KL) divergence between the output distributions of \( \Theta_u \) and \( \Theta \) on \( D_r \):

\vspace{-6mm}
\begin{equation}
\small
\mathcal{L}_{\text{CM}} = \frac{1}{N_r} \sum_{n=1}^{N_r} \Big[ 
\text{KL}\left( \mathbf{p}_o^{\text{img}} \parallel \mathbf{p}_u^{\text{img}} \right) 
+ \text{KL}\left( \mathbf{p}_o^{\text{txt}}\parallel \mathbf{p}_u^{\text{txt}}\right) \Big]
\label{eq:cm_loss}
\end{equation}
where \( \mathbf{p}_o^{\text{img}} \) and \( \mathbf{p}_u^{\text{img}} \) are the probability distributions derived from the image embeddings produced by the original model \( \Theta \) and the unlearned model \( \Theta_u \) after the softmax, respectively. Similarly, \( \mathbf{p}_o^{\text{txt}} \) and \( \mathbf{p}_u^{\text{txt}} \) represent the distributions derived from the text embeddings of \( \Theta \)  and \( \Theta_u \), respectively. 



Combing these three modules, our overall unlearning loss is:
\begin{equation}
\small
\mathcal{L} = \lambda_1\mathcal{L}_{\text{RM}} + \lambda_2 \mathcal{L}_{\text{FM}} + \lambda_3 \mathcal{L}_{\text{CM}}
\label{eq:total_loss}
\end{equation} 
where \( \lambda_1 \), \( \lambda_2 \), and \( \lambda_3 \) are hyperparameters  control the relative importance of each module in the overall unlearning process. These hyperparameters allow us to fine-tune the balance between forgetting the information in \(D_f\), retaining performance on \(D_r\), and ensuring consistency between the original and unlearned models. 


\section{Experimental Evaluation}

\subsection{Experiments Setting}

Table \ref{setting1} summarizes our experimental settings, including tasks, datasets, and evaluation metrics.  We refer readers to Appendix~\ref{setting} for more experiments setting details.

\noindent
\textbf{Tasks:} To evaluate the effectiveness of our method, we conducted experiments on five tasks:

\noindent
\textit{\underline{1. Zero-shot Image Prediction:}} 
CLIP predicts images by comparing their embeddings with text embeddings of class names and selecting the most similar one. The unlearned model should misclassify images from $D_f$, demonstrating its forgetting ability.   

\noindent
\textit{\underline{2. Zero-shot Text Retrieval:}} 
Given an image, CLIP retrieves the most relevant text from a predefined set of class names or concepts based on similarity scores, evaluating its zero-shot semantic alignment.  

\noindent
\textit{\underline{3. Image Retrieval (IR):}} 
Given a text query, CLIP retrieves images by ranking them based on similarity to the text embedding.

\noindent
\textit{\underline{4. Text Retrieval (TR):}}
Given an image, CLIP retrieves text descriptions by ranking them based on similarity to the image embedding.  

\noindent
\textit{\underline{5.  Diffusion-based Image Generation:}} Using Stable Diffusion~\cite{rombach2022high}, CLIP's text encoder should fail to generate images containing forgotten content while preserving accuracy for retained elements.

\begin{table}[t]
\renewcommand{\arraystretch}{1.0}
\scalebox{0.56}{
\begin{tabular}{c|c|cc|cc}
\toprule
\textbf{Dataset} & \textbf{Method} & \multicolumn{2}{c|}{\textbf{ZS Prediction (\%)}} & \multicolumn{2}{c}{\textbf{ZS Retrieval (\%)}} \\
\cmidrule(lr){3-4} \cmidrule(lr){5-6}
                  &                 & Acc.$D_f$ $\downarrow$ & Acc.$D_r$ $\uparrow$ & Acc.$D_f$ $\downarrow$ & Acc.$D_r$ $\uparrow$ \\
\midrule
\multirow{6}{*}{\textbf{CIFAR-100}} 
                  & CLIP            & 86.08  & 72.85  & 88.61 & 73.43  \\
                  & CLIP+GA         & 4.43  & 5.22  & 0.63  & 5.39 \\
                  & CLIP+GradDiff   & 0.00   & 89.96  & 0.00  & 90.64  \\
                  & CLIP+KL         & 91.88   & 80.88  & 91.77  & 81.51  \\
                  & CLIP+ENMN       & 0.00   & 12.46  & 0.00  & 17.94  \\
                  & \cellcolor{gray!20}CLIPErase (ours) & \cellcolor{gray!20}0.00 & \cellcolor{gray!20}90.99 & \cellcolor{gray!20}0.00 & \cellcolor{gray!20}91.85 \\
\midrule
\multirow{6}{*}{\textbf{Conceptual 12M}} 
                  & CLIP            & 96.20 & 93.60 & 94.48 & 92.77 \\
                  & CLIP+GA         & 38.22  & 4.15   & 1.17  & 5.38   \\
                  & CLIP+GradDiff   & 4.96   & 97.01  & 5.64   & 97.46  \\
                  & CLIP+KL         & 99.04  & 98.41  & 98.83  & 98.02  \\
                  & \cellcolor{gray!20}CLIPErase (ours) & \cellcolor{gray!20}0.74 & \cellcolor{gray!20}97.10 & \cellcolor{gray!20}0.74 & \cellcolor{gray!20}97.62 \\
\bottomrule
\end{tabular}
}
\caption{Experiment results on CIFAR-100 and Conceptual 12M datasets for Zero-shot (ZS) Prediction and Retrieval tasks.}
\label{vertical}
\end{table}

\noindent
\textbf{Implementation Details:} 
In our experiments, we use three datasets:  CIFAR-100~\cite{krizhevsky2009learning}, Conceptual 12M~\cite{changpinyo2021conceptual} and Flickr30K~\cite{bojchevski2017deep}.
For CIFAR-100 and Conceptual 12M, we randomly select one or more classes as the forget set. The model is trained for 20 epochs with a batch size of 16, using the Adam optimizer and an initial learning rate of $1 \times 10^{-6}$.
For Flickr30K, the forget set consists of all image-text pairs containing a specific concept (approximately 1\% of the dataset). The same training setup is used, except with a lower learning rate of $1 \times 10^{-8}$. All datasets are split into 70\% training and 30\% testing.  The balance hyperparameters are set as $\lambda_1 = 1$, $\lambda_2 = \lambda_3 = 3$.
Experiments use the best checkpoint from the validation set. Our model is implemented in PyTorch and trained on an NVIDIA V100. 

\noindent 
\textbf{Metrics:} Following prior work~\cite{Kim2021DiffusionCLIPTD}, we use two metrics to assess MU efficacy: (1) Retain Set Performance ($D_r$ $\uparrow$): evaluates the accuracy for Zero-shot tasks and the Image Diffusion task, as well as recall@1, @5, and @10 for Retrieval tasks.  Higher values indicate minimal impact on the retain set after unlearning. (2) Forget Set Performance ($D_f$ $\downarrow$): evaluates the same accuracy with (1) but in forget set $D_f$. Lower values indicate more effective unlearning.

\noindent 
\textbf{Baselines:} 
Besides the original CLIP model~\cite{Kim2021DiffusionCLIPTD}, we include the following unlearning methods for comparison:

\noindent
\textit{\underline{1. Gradient Ascent (GA)}}~\cite{yao2023large}: Increases prediction errors on the forget set, forcing the model away from its original predictions.

\noindent
\textit{\underline{2. Gradient Difference (GradDiff)}}~\cite{liu2022continual}: Increases errors on the forget set while preserving performance on the retain set.

\noindent
\textit{\underline{3. KL Minimization (KL)}}~\cite{maini2024tofu}: Aligns prediction distributions between unlearned and original models for the retain set while increasing errors on the forget set.

\noindent
\textit{\underline{4. Error Minimization-Maximization Noise (EMMN)}} \cite{chundawat2023zero}: Removes class information using noise-based error minimization and maximization techniques.

\subsection{Zero-shot Prediction and Retrieval}

As shown in Table~\ref{vertical}, 
CLIPErase preserves CLIP’s robust performance and achieves superior results after accurately deleting specific information.
On the Forget Set, it attains 0\% accuracy in both tasks, confirming complete unlearning.
CLIP+GA, CLIP+GradDiff, and CLIP+KL leave residual accuracy, while CLIP+ENMN also reaches 0\%, but drastically reduces retention performance to 12.46\% in prediction and 17.94\% in retrieval.
In contrast, CLIPErase excels on the Retain Set with 90.99\% in prediction and 91.85\% in retrieval, an improvement of 18.14\% and 18.42\% over the original CLIP.
This success arises from CLIPErase’s Retention and Consistency Modules, which safeguard knowledge of retained data and enhance alignment between visual and textual features.

To achieve finer-grained unlearning on larger, open-domain datasets, we test CLIPErase on Conceptual 12M. Table \ref{vertical} shows that CLIPErase reduces the Forget Set accuracy to 0.3\% in prediction and 0.4\% in retrieval, while preserving strong performance on the Retain Set at 94.78\% in prediction and 94.81\% in retrieval. These findings confirm CLIPErase’s ability to generalize to open-domain data, deliver precise unlearning for the Forget Set, and maintain robust performance on the Retain Set, underscoring its potential for broader applications across diverse datasets. In Appendix~\ref{Performance Stability and Variance}, we analyze forget set classes by computing mean and variance, confirming CLIPErase’s consistent unlearning.

\begin{table}[t]
\renewcommand{\arraystretch}{1.0}  
\scalebox{0.54}{  
\begin{tabular}{c|c|cc|cc|cc}
\toprule
\textbf{Task} & \textbf{Method} & \multicolumn{2}{c|}{R@1 (\%)} & \multicolumn{2}{c|}{R@5 (\%)} & \multicolumn{2}{c}{R@10 (\%)} \\ 
\cmidrule(lr){3-4} \cmidrule(lr){5-6} \cmidrule(lr){7-8}
              &                 & $D_f$ $\downarrow$  & $D_r$ $\uparrow$  & $D_f$ $\downarrow$  & $D_r$ $\uparrow$  & $D_f$ $\downarrow$  & $D_r$ $\uparrow$  \\ 
\midrule
\multirow{6}{*}{\textbf{Image Retrieval}} 
              & CLIP            & 28.61 & 22.76 & 56.75 & 50.14 & 66.73 & 60.67 \\ 
              & CLIP+GA         & 0.00 & 0.00 & 0.00 & 0.00 & 0.00 & 0.00 \\ 
              & CLIP+GradDiff   & 2.67 & 16.26 & 6.21 & 37.49 & 7.82 & 47.13 \\ 
              & CLIP+KL         & 26.30 & 22.94 & 53.63 & 48.56 & 63.17 & 58.52 \\ 
              & CLIP+ENMN       & 12.54 & 19.00 & 24.81 & 33.23 & 49.72 & 38.64 \\ 
              & \cellcolor{gray!20}CLIPErase (ours) & 
                \cellcolor{gray!20}3.37 & \cellcolor{gray!20}17.71 & 
                \cellcolor{gray!20}8.36 & \cellcolor{gray!20}40.24 & 
                \cellcolor{gray!20}10.55 & \cellcolor{gray!20}50.35 \\
\midrule
\multirow{6}{*}{\textbf{Text Retrieval}} 
              & CLIP            & 25.04 & 19.11 & 58.87 & 48.82 & 68.33 & 59.17 \\ 
              & CLIP+GA         & 0.00 & 0.00 & 0.00 & 0.00 & 0.00 & 0.01 \\ 
              & CLIP+GradDiff   & 2.05 & 14.10 & 6.05 & 37.58 & 7.28 & 47.19 \\ 
              & CLIP+KL         & 19.91 & 17.17 & 49.81 & 44.85 & 59.52 & 54.85 \\ 
              & CLIP+ENMN       & 13.53 & 14.90 & 24.50 & 30.01 & 29.31 & 35.66 \\ 
              & \cellcolor{gray!20}CLIPErase (ours) & 
                \cellcolor{gray!20}2.35 & \cellcolor{gray!20}13.82 & 
                \cellcolor{gray!20}7.21 & \cellcolor{gray!20}37.06 & 
                \cellcolor{gray!20}8.84 & \cellcolor{gray!20}46.52 \\
\bottomrule
\end{tabular}
}
\caption{Experiment results on the Flickr30K dataset for Retrieval tasks. }
\label{cmp_retrieval}
\end{table}

\subsection{ Image Retrieval and Text Retrieval }

As shown in Table~\ref{cmp_retrieval},
CLIPErase delivers strong performance in both image and text retrieval tasks for multimodal unlearning, surpassing prior methods on the Flickr30K dataset.
In image retrieval, CLIPErase significantly outperforms CLIP+GA, CLIP+GradDiff, and CLIP+KL, with notable gains in R@1 (17.71\%), R@5 (40.24\%), and R@10 (50.35\%).
Similarly, it achieves competitive results in text retrieval, reaching 46.52\% in R@10, which is 8.33\% higher than CLIP+KL.

These results highlight the effectiveness of CLIPErase’s Forgetting, Retention, and Consistency Modules. The Forgetting Module successfully reduces model reliance on forgotten data, while the Retention and Consistency Modules preserve performance on the retain set, ensuring robust multimodal unlearning without retraining.

\subsection{Ablation Studies}

As shown in Table \ref{ablation}, we performed ablation studies on the CIFAR-100 dataset to quantitatively evaluate the impact of the Forgetting Module (FM), Retention Module (RM), and Consistency Module (CM) on the zero-shot prediction task.

\begin{table}[t]
\centering
\footnotesize
\renewcommand{\arraystretch}{1}
\scalebox{0.97}{
\begin{tabular}{ccc|cc|cc}
\toprule
\multirow{2}{*}{FM} & \multirow{2}{*}{RM} & \multirow{2}{*}{CM} & \multicolumn{2}{c|}{Accuracy (\%)} & \multicolumn{2}{c}{Improvement (\%)} \\
\cmidrule(lr){4-5} \cmidrule(lr){6-7}
  &   &   & $\downarrow D_f$ & $\uparrow D_r$ & $\downarrow D_f$ & $\uparrow D_r$ \\
\midrule 
\xmark & \xmark & \xmark & 86.08 & 72.85 & - & - \\
\cmark & \xmark & \xmark & 18.57 & 64.12 & $\downarrow 67.5$ & $\downarrow 8.73$ \\
\cmark & \cmark & \xmark & 9.40  & 73.14 & $\downarrow 76.68$ & $\uparrow 0.56$ \\
\cmark & \cmark & \cmark & 0     & 	90.80 & $\downarrow 86.08$ & $\uparrow 17.95$ \\
\bottomrule
\end{tabular}
}
\caption{Ablation studies on the Forgetting Module (FM), Retention Module (RM), and Consistency Module (CM).}
\label{ablation}
\end{table}

\noindent
\textbf{Effectiveness of FM:} Activating the Forgetting Module led to a significant drop in accuracy on the forget set, from 86.08\% to 18.57\%, indicating that FM effectively disrupted the correspondence between images and texts in the forget set. However, relying solely on FM negatively impacted the retain set performance, reducing its accuracy from 72.85\% to 64.12\%.

\noindent
\textbf{Effectiveness of RM:} When both the Forgetting Module and Retention Module were activated, the accuracy on the retain set recovered to 73.14\%, demonstrating that RM successfully protected the retain set's performance. Simultaneously, the model’s accuracy on the forget set further decreased to 9.40\%, showing that RM plays a critical role in balancing the task of forgetting specific data while preserving the retain set performance.

\noindent
\textbf{Effectiveness of CM:} When the Consistency Module (CM) was activated alongside FM and RM, the retain set accuracy significantly improved to 90.80\%, while the accuracy on the forget set dropped to 0\%. This indicates that the model successfully forgot the specified data while maintaining high performance on the retain set. Consistency Module  ensures consistency throughout the unlearning process, preventing potential errors or biases introduced by the unlearning process.

\begin{figure}[t]
\centering
\includegraphics[width=0.5\textwidth]{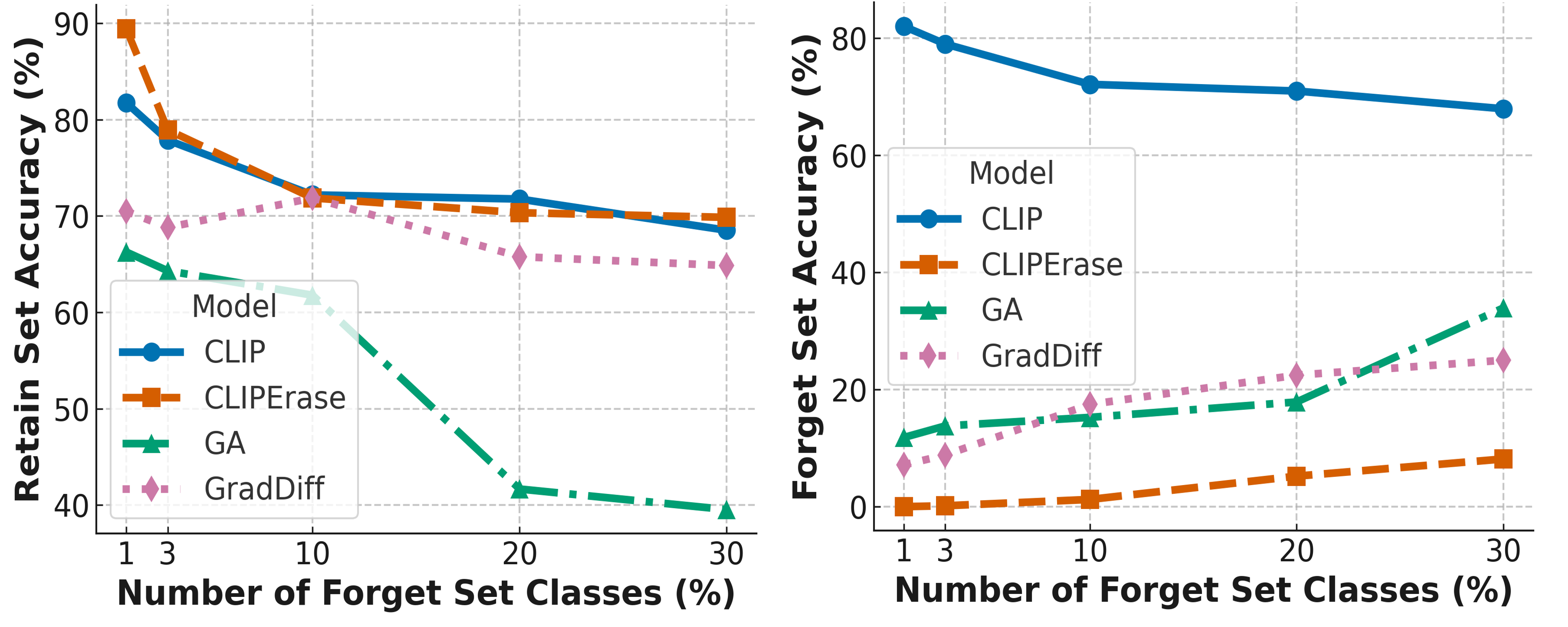}
\caption{Performance across different numbers of Forget Set classes. 
}
\label{comparison}
\end{figure}

\subsection{Robustness and Utility}

Figure~\ref{comparison} shows experiments with CLIPErase, GradDiff, GA, and the original CLIP across various proportions of forget classes (0\%, 3\%, 10\%, 20\%, 30\%) on the CIFAR-100 dataset for the zero-shot image classification task.   Figure~\ref{comparison} 
 (left)  shows the forget accuracy, where CLIPErase demonstrates a significant unlearning effect at different proportions of forget classes. However, for GradDiff and GA, forget accuracy increases as the number of forget classes grows, indicating a worsening unlearning effect. Figure~\ref{comparison} (right)   presents the retain accuracy, where CLIPErase maintains performance comparable to CLIP, particularly at lower proportions, while GA shows a sharp decline in retain accuracy as the forget proportion increases. Overall, CLIPErase exhibits strong robustness and utility against the number of forget classes, effectively balancing both forget accuracy and retain accuracy. 
We further validate CLIPErase’s robustness with experiments on Conceptual 12M, detailed in Appendix~\ref{Scalabilit}.

\subsection{Extending CLIPErase to Other VLMs}

Although our main study focuses on CLIP, CLIPErase is designed to be modular and model-agnostic, without relying on any CLIP-specific components. To demonstrate its extensibility, we further applied CLIPErase to other vision-language models, including BLIP~\cite{li2022blip,li2023blip}, ALBEF~\cite{li2021align}, and other transformer-based architectures.

To validate this generalizability, we implemented CLIPErase on BLIP-1~\cite{li2022blip} and conducted additional experiments on the CIFAR-100 dataset for the zero-shot prediction task shown on Table~\ref{tab:blip}. We applied our Forgetting, Retention, and Consistency Modules without requiring any changes to BLIP's original architecture.

\begin{table}[t]
\centering
\renewcommand{\arraystretch}{1.0}
\scalebox{0.78}{
\begin{tabular}{l|cc}
\toprule
Model & Acc.$D_f$ $\downarrow$ & Acc.$D_r$ $\uparrow$  \\
\midrule
BLIP & 100.00 & 97.07 \\
BLIP + GA & 0.00 & 42.89 \\
BLIP + GradDiff & 89.73 & 40.41 \\
\cellcolor{gray!20}BLIP + CLIPErase & \cellcolor{gray!20}0.00 & \cellcolor{gray!20}83.12 \\
\bottomrule
\end{tabular}
}
\caption{Experiment results on CIFAR-100 for Zero-shot (ZS) Prediction task using BLIP.}
\label{tab:blip}
\end{table}

We further evaluated the effectiveness of CLIPErase on BLIP and compared it with baseline methods, as shown in Table~\ref{tab:blip}. The original BLIP model retains all information in the forget set with an accuracy of 100.00\% and achieves 97.07\% on the retain set, indicating no unlearning effect. While BLIP + GA and BLIP + GradDiff reduce forget-set accuracy to 0.00\% and 89.73\% respectively, they also lead to a substantial drop in retain-set performance, with accuracies of 42.89\% and 40.41\%. In contrast, CLIPErase reduces the forget-set accuracy to 0.00\% and maintains a strong retain set accuracy of 83.12\%. These results demonstrate that CLIPErase enables effective unlearning on BLIP with minimal impact on retained knowledge, highlighting its generalizability beyond CLIP.

\section{Diffusion Model with CLIPErase}

To further demonstrate the practical impact of CLIPErase and its ability to remove specific associations while preserving other knowledge, we apply it to a text-to-image diffusion model~\cite{rombach2022high}. Our goal is to selectively erase targeted object concepts from generated images while maintaining other details. We conduct this experiment using captions from the Flickr30k dataset, which contains real-world images and complex textual descriptions, providing a challenging testbed for evaluating the precision of our unlearning approach.
For instance, consider the caption:  \emph{“A woman holding an apple standing next to a display of oranges, apples, and melons.”}  This caption includes multiple coexisting concepts (\emph{woman}, \emph{apple}, \emph{oranges}, \emph{melons}). If \emph{apple} is designated as the target for removal, we assess whether CLIPErase can effectively erase it from the generated image while ensuring that other elements, such as the \emph{woman}, \emph{oranges}, and \emph{melons}, remain present.

\begin{figure}[t]
\centering
\includegraphics[width=0.5\textwidth]{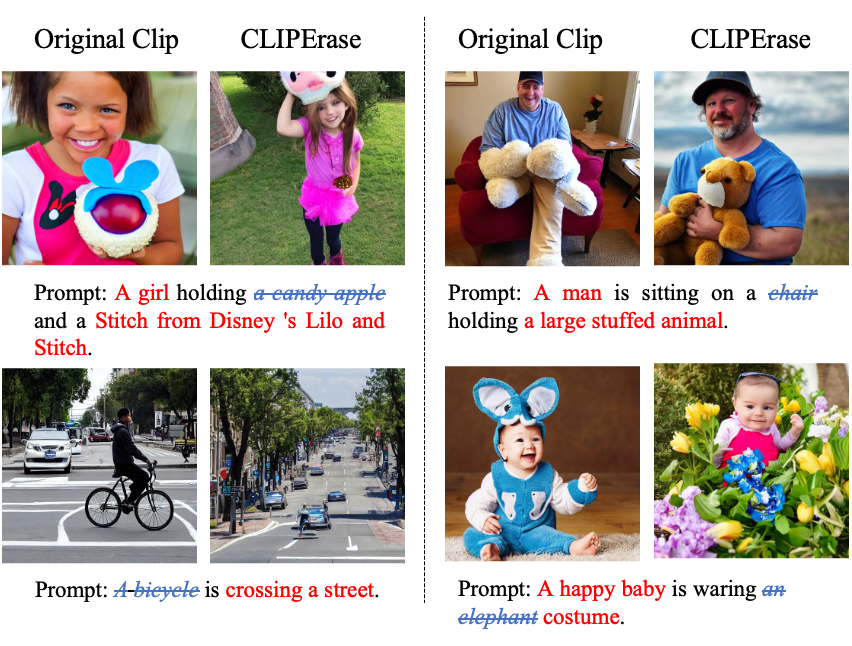}
\caption{Comparison of image generation results using the original CLIP and our CLIPErase model in Stable Diffusion with multi-concept prompts. The prompt represents the input to the diffusion model. Blue text denotes concepts unlearned by CLIPErase, while red text highlights concepts that should be retained.
}
\label{diffu}
\end{figure}

\begin{table}[t]
\centering
\small
\begin{tabular}{l|cc}
\toprule
Unlearned Concept & CLIP(\%) & CLIPerase (\%) \\
\midrule
Apple     & 100.00  & 2.00 \\
Bicycle   & 90.00  & 8.00 \\
Chair     & 84.00  & 6.00 \\
Elephant  & 88.00  & 6.00 \\
\bottomrule
\end{tabular}
\caption{Detection rate (\%) of different unlearned concepts in generated images. Lower values indicate more effective removal.} 
\label{table:diffusion_results}
\end{table}

We generate images using two models: (1) one with the standard CLIP text encoder and (2) another with the CLIP text encoder modified by CLIPErase.  We select 50 captions per target concept and generate 400 images per model. For evaluation, we then use a pretrained YOLOv5~\cite{zhang2022real} detector to identify the presence of the target object in the generated images. The detection rate serves as a metric for evaluating concept removal effectiveness: \( \text{Detection Rate} = \frac{N_d}{N_g} \), where \( N_d \) is the number of images where the target concept is detected, and \( N_g \) is the total number of generated images. A lower detection rate indicates more effective concept removal.

As shown in Figure~\ref{diffu},  
CLIPErase effectively removes specific target concepts while preserving other relevant concepts in the generated images. Table~\ref{table:diffusion_results} shows that CLIPerase significantly reduces target concept presence. For example, the detection rate for apple drops from 100.00\% (standard CLIP) to 2.00\% (CLIPerase), and bicycle from 90.00\% to 8.00\%, demonstrating successful selective unlearning. We present additional results in Appendix~\ref{diffusion_more}.

\section{Visualization} 

\begin{figure}[t]
\centering
\includegraphics[width=0.45\textwidth]{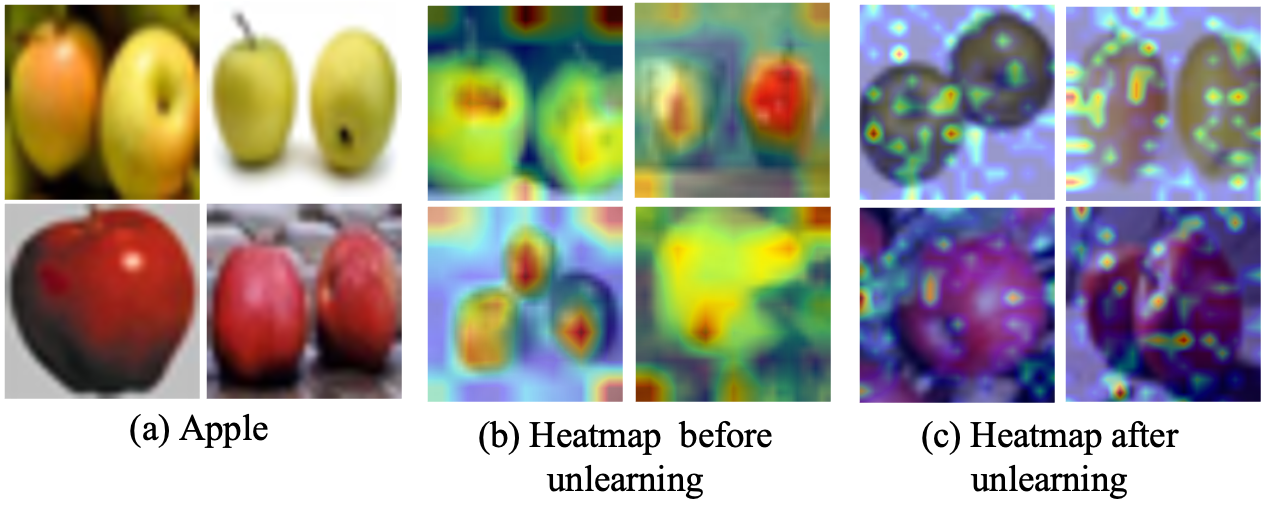}
\caption{
Attention Heatmaps before unlearning (CLIP) and after unlearning (CLIPErase) on apple images.
}

\label{heatmap}
\end{figure}

\noindent
\textbf{Attention Visualization:} In Fig.~\ref{heatmap}, we visualized the attention heatmaps of the "forget set" before and after the unlearning process. Using "apple" from CIFAR-100  as an example, we presented the original images, the heatmaps generated by the CLIP model, and the heatmaps after unlearning with CLIPErase. In the CLIP heatmaps, attention is highly focused on the object, displaying strong visual-semantic alignment. This indicates that CLIP successfully establishes a robust connection between textual and visual semantics. In contrast, after machine unlearning with CLIPErase, the heatmaps show that attention becomes more random and dispersed across each patch, no longer concentrated on the relevant object. This suggests that CLIPErase effectively disrupts the alignment between text and visual semantics for the data to be unlearned, thus achieving the intended unlearning objective.

\begin{figure}[t]
\centering
\includegraphics[width=0.47\textwidth]{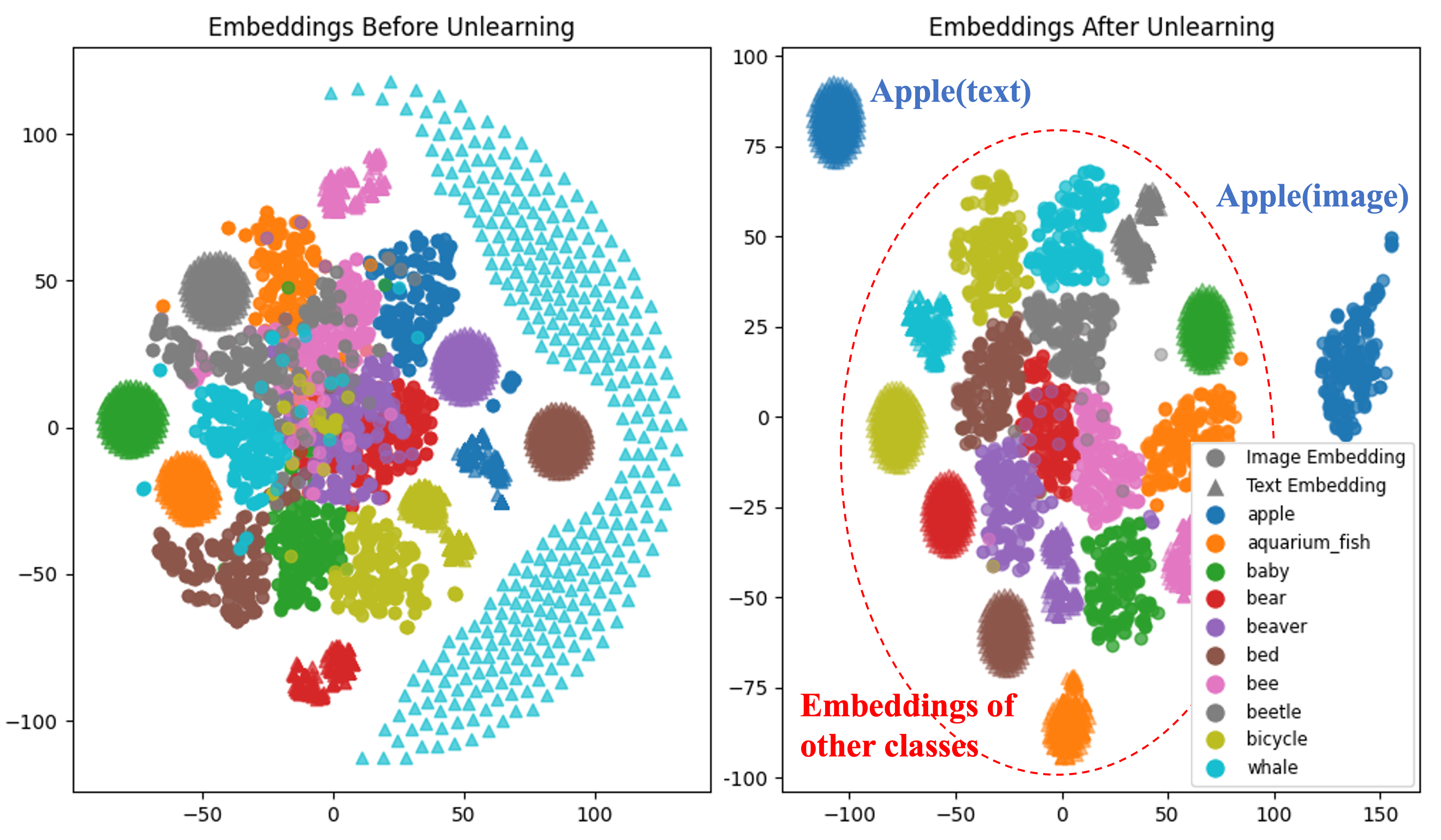}
\caption{t-SNE visualizations of text and visual embeddings from the CLIP model (before unlearning, left) and the CLIPErase model (after unlearning, right) on CIFAR-100. The unlearned category is "apple." We use $\circ$ and $\triangle$ to represent visual and text modalities, respectively, with different colors indicating different categories. 
}
\label{tsne}
\end{figure}

\noindent
\textbf{Embedding Visualization:} To further investigate the impact of CLIPErase on the retain set within the cross-modal shared representation space, we conducted t-SNE visualizations \cite{van2008visualizing} of the textual and visual embeddings from both the original CLIP model and the CLIPErase model. Specifically, we selected 10 classes from the CIFAR-100 dataset, with the "apple" class serving as the forget set and the remaining classes as the retain set. 

From the CLIPErase results, we observed that the distance between the textual and visual embeddings of the "apple" class significantly increased, while the embeddings of the other classes remained tightly clustered within their respective categories. This suggests that CLIPErase effectively weakens the association between the textual and visual modalities in the forget set, with minimal impact on the modality associations of the retain set. In summary, CLIPErase can effectively decouple the connection between the textual and visual embeddings of the forget set without affecting the visual-textual associations of the retain set, thus achieving the intended goal of machine unlearning.

\section{Conclusion}

In this paper, we introduce CLIPErase, a novel machine unlearning framework for multimodal models that selectively removes undesired associations while preserving overall model performance. CLIPErase achieves this by disrupting cross-modal associations for the specified data, effectively "forgetting" the targeted information without compromising the model's ability to perform other tasks. The effectiveness of CLIPErase is demonstrated through extensive experiments on various tasks, including zero-shot prediction, image-text retrieval, and text-to-image generation with diffusion models.  The results consistently show that CLIPErase successfully removes targeted associations while maintaining performance on other tasks and data. This highlights CLIPErase's potential for addressing real-world challenges related to privacy preservation, intellectual property protection, and bias mitigation in multimodal learning.


\section*{Acknowledgement}

The authors would like to thank Zheyuan Liu and Professor Meng Jiang from the University of Notre Dame for their valuable feedback and discussions on early versions of this work. This work was supported by NSF Award No. 2333795.


\section*{Limitation}
A key limitation of CLIPErase is the lack of dedicated datasets and benchmarks designed for multimodal machine unlearning. Existing benchmarks are not explicitly constructed to evaluate the effectiveness of multimodal unlearning, limiting comprehensive assessment. Moreover, our work only focus on unlearning multimodel embedding models. We plan to extend our framework to multimodal generative models such as Vision Language Models (VLMs), enabling more direct and effective unlearning of visual-textual associations for privacy preservation, intellectual property protection, and bias mitigation. Further discussions on future directions are provided in Appendix~\ref{future}.

\bibliography{custom}

\appendix

\appendix
\section*{Appendix}
\addcontentsline{toc}{section}{Appendix} 

In these supplementary materials, we provide details on our experimental settings in Appendix~\ref{setting}, and different patch size model results in Appendix~\ref{patch14}. We further analyze variance effects in Appendix~\ref{Variance}, scalability in Appendix~\ref{Scalabilit}, additional diffusion model results in Appendix~\ref{diffusion_more}, and future directions in Appendix~\ref{future}.

\section{Experiments Setting}
\label{setting}

\subsection{Tasks:}

\noindent
\textit{\underline{1. Zero-shot Prediction:}} CLIP is employed to classify images based on a given category label (e.g., "cat," "dog") without training data. The unlearned CLIP is expected to predict a mismatch between the label ``dog'' and the image of a dog if it belongs to the forget set ($D_f$), demonstrating its ability to "forget" specific learned associations.

\noindent
\textit{\underline{2. Zero-Shot Prediction:}} CLIP is used to retrieve images matching a text description (e.g., "red car") without task-specific training. In the unlearning scenario, the unlearned CLIP should fail to retrieve images that belong to the forget set, such as cars described as "red car" if they are part of $D_f$, demonstrating its ability to forget these specific multimodal associations without further training.

\noindent
\textit{\underline{3. Image Retrieval (IR):}} Given a text query, CLIP retrieves the top-$k$ relevant images. The unlearned CLIP should avoid retrieving images in the forget set, demonstrating the selective forgetting capability while maintaining performance on the retain set.

\noindent
\textit{\underline{4. Text Retrieval (TR):}} Given an image query, CLIP retrieves the top-$k$ relevant text descriptions. In the unlearning context, CLIP should fail to retrieve relevant text descriptions for images in the forget set.

\noindent \textit{\underline{5. Image Diffusion:}} We utilize Stable Diffusion, whose text encoder is based on CLIP, to generate images from textual prompts. Under the unlearning scenario, this CLIP-based encoder is expected to "forget" all content in the forget set ($D_f$). Consequently, when given a prompt that references any forgotten content, the unlearned model should fail to produce images containing those elements, thus illustrating the selective forgetting capability while still maintaining accurate generation for all other retained content.

\subsection{Datasets:}
For the zero-shot prediction and retrieval tasks, we use the CIFAR-100 dataset~\cite{krizhevsky2009learning}. CIFAR-100 is an image classification dataset containing 100 categories. It consists of a total of 60,000 images, with 50,000 used for training and 10,000 for testing. 

Similarly, we also extend our experiments to the Conceptual 12M dataset~\cite{changpinyo2021conceptual}, a large-scale, open-domain dataset comprising approximately 12 million image-text pairs sourced from the web. Conceptual 12M provides diverse and natural text descriptions, making it particularly suitable for training multimodal models across various tasks, which allows us to explore whether the CLIPErase can generalize to larger class sets or more diverse real-world data.

For the Image Retrieval (IR) and Text Retrieval (TR) tasks, we use the Flickr30K dataset~\cite{bojchevski2017deep}. The dataset contains 31,783 images, each paired with five natural language captions. These images primarily depict daily life scenes. Additionally, for the Image Diffusion task, we generate images using the captions provided in Flickr30K.

\subsection{Settings:} 
For CIFAR-100, we randomly select one or multiple classes as the forget set. The model is trained for 20 epochs with a batch size of 16, using Adam and an initial learning rate of $1 \times 10^{-6}$. The loss weights are set to $\lambda_1 = 1, \lambda_2 = \lambda_3 = 3$. 
 
For CC12M, although it contains 12 million real-world text-image pairs, we faced practical challenges such as broken image links and the extensive time required for data collection. On our servers, the download and cleaning pipeline processes approximately 1,000 text-image pairs per hour, implying that collecting the full dataset would take over 12,000 hours. Due to these constraints, we randomly sampled 120,000 vision-language pairs (1\% of CC12M), comprising a Forget Set with 2,284 images and a Retain Set with 117,716 images. For this experiment, we designated the concepts \textit{woman} and \textit{womens} as the forget targets, and performed Zero-Shot Image and Text Retrieval tasks. The model was trained for 5 epochs with a batch size of 16, using a 70\%-30\% train-validation split. We set the learning rate to \(1 \times 10^{-6}\), weight decay to \(1 \times 10^{-5}\), and loss weights to \(\lambda_1 = 1\), \(\lambda_2 = \lambda_3 = 3\).

For Flickr30k, the forget set includes all image-text pairs containing one concept, comprising 1\% of the dataset. The same setup is used, except with an initial learning rate of $1 \times 10^{-8}$. The loss weights are set to $\lambda_1 = 1, \lambda_2 = \lambda_3 = 3$. 

Each experiment use the best checkpoint from the validation set. Our model and code are implemented in PyTorch, with training and evaluation on an NVIDIA Tesla V100-SXM2.

\subsection{Evaluation:} 
Following prior work~\cite{Kim2021DiffusionCLIPTD}, we use two metrics to assess MU efficacy: (1) Retain Set Performance ($D_r$ $\uparrow$):This metric evaluates the accuracy for Zero-shot tasks and the Image Diffusion task, as well as recall@1, @5, and @10 for Retrieval tasks.  Higher values indicate minimal impact on the retain set after unlearning. (2) Forget Set Performance ($D_f$ $\downarrow$): Uses the same metrics as $D_r$. Lower values indicate more effective unlearning.

\subsection{Comparison to Prior Work}
\label{Comparison to Prior Work}

Besides the original CLIP model, we compare our method CLIPErase with commonly known unimodal MU methods when directly applied to CLIP. We omit using prior work~\cite{kravets2024zero} as a baseline because their code and detailed experimental settings were not released, making it challenging to replicate their results. Especially without access to their synthetic image generation code, it is unable to conduct CLIP unlearning in their setting.

\noindent
\textit{\underline{Original CLIP Model}}~\cite{Kim2021DiffusionCLIPTD}: We use the original CLIP model as a baseline to assess the impact of unlearning, ensuring that the model’s performance on the retain set remains unaffected.

\noindent
\textit{\underline{Gradient Ascent (GA)}}~\cite{yao2023large}: This method aims to degrade the model's performance on the forget set by increasing prediction errors, forcing the model behave away from its original predictions.

\noindent
\textit{\underline{Gradient Difference (GradDiff)}}~\cite{liu2022continual}: This method increases errors on the forgetting data while preserving performance on the retain set, achieving the goal of unlearning specific information without impacting retained data.

\noindent
\textit{\underline{KL Minimization (KL)}}~\cite{maini2024tofu}: The method ensures consistency on the retain set by comparing the prediction distributions of the unlearned and original models, while increasing errors on the forgetting data.

\noindent
\textit{\underline{Error Minimization-Maximization Noise (EMMN)}} \cite{chundawat2023zero}: It deletes specific class information from the model using noise-based error minimization and maximization techniques.

\subsection{Computational Efficiency}
\label{efficiency}

We also evaluated the computational burden and scalability of our method, CLIPErase. The average training time per epoch is approximately 39 minutes for CIFAR-100, 98.7 minutes for Conceptual 12M, and 146.88 minutes for Flickr30K on an NVIDIA Tesla V100-SXM2 GPU. Since we did not profile the time consumption across different components of the training pipeline, there may be further opportunities to reduce overhead and improve efficiency.

\section{Performance Stability and Variance}
\label{Performance Stability and Variance}

To ensure that CLIPErase consistently performs effective unlearning across various forget sets, it is essential to evaluate not only the average performance but also the variability of the results. To achieve this, we conducted experiments and calculated the mean and variance. A low variance indicates that the method performs uniformly well across different forget sets, while a high variance may suggest sensitivity to specific classes.

We conducted experiments on the CIFAR-100 dataset using five randomly selected forget classes: Apple, Camel, Mountain, Porcupine, and Television. For each forget class, we measured the performance of CLIPErase in Zero-Shot Text Retrieval and Zero-Shot Prediction tasks. The performance metrics, including the mean and variance of the results, are summarized in Table~\ref{tab:stability_variance_results}.

\begin{table}[ht]
\centering
\footnotesize
\renewcommand{\arraystretch}{1}
\scalebox{1}{
\begin{tabular}{l|cc|cc}
\toprule
\textbf{Class} & \multicolumn{2}{c|}{\textbf{ZS Retrieval (\%)}} & \multicolumn{2}{c}{\textbf{ZS Prediction (\%)}} \\
\cmidrule(lr){2-3} \cmidrule(lr){4-5}
               & \(D_f\) & \(D_r\) & \(D_f\) & \(D_r\) \\
\midrule 
Apple         & 0.00       & 89.41      & 0.00       & 91.38      \\
Camel         & 0.00       & 80.61      & 0.00       & 78.36      \\
Mountain      & 8.19       & 80.81      & 0.44       & 78.60      \\
Porcupine     & 0.00       & 80.28      & 0.00       & 77.71      \\
Television    & 0.41       & 81.72      & 0.00       & 78.41      \\
\midrule
\textbf{Mean}      & \textbf{1.72}  & \textbf{82.57} & \textbf{0.09} & \textbf{80.89} \\
\textbf{Variance}  & \textbf{13.11} & \textbf{14.92} & \textbf{0.04}  & \textbf{34.48}  \\
\bottomrule
\end{tabular}
}
\caption{Performance of CLIPErase on different forget classes in Zero-Shot Prediction and  Retrieval tasks. Metrics are reported for Forget Set (\(D_f\)) and Retain Set (\(D_r\)).}
\label{tab:stability_variance_results}
\end{table}

The experimental results demonstrate that Zero-Shot Prediction exhibits the most stable forgetting performance, as indicated by the low variance in the forget set performance (\(0.038\%\)). This suggests that CLIPErase consistently forgets across various classes in this task. In contrast, Zero-Shot Text Retrieval shows a higher variance (\(13.11\%\)) for the forget set, which reflects the inherent complexity and variability of text-image associations in certain classes like Mountain. Despite this variability, the overall mean performance remains low (\(1.72\%\)), indicating effective forgetting.

Furthermore, the retain set performance maintains relatively low variance for both tasks (\(14.92\%\) for Text Retrieval and \(34.48\%\) for Image Retrieval), demonstrating that the model consistently retains unrelated concepts across different forget sets. This robustness highlights CLIPErase’s ability to perform unlearning reliably without compromising the retention of non-forget classes.

Overall, the variance analysis confirms that while different forget sets can influence unlearning performance to some extent, CLIPErase maintains strong stability and low variance across varied forget sets. 

\subsection{More results of different Patch size}
\label{patch14}

\begin{table}[ht]
\centering
\footnotesize
\renewcommand{\arraystretch}{1}
\scalebox{1}{
\begin{tabular}{l|cc|cc}
\toprule
\textbf{Class} & \multicolumn{2}{c|}{\textbf{ZS Prediction (\%)}} & \multicolumn{2}{c}{\textbf{ZS  Retrieval (\%)}} \\
\cmidrule(lr){2-3} \cmidrule(lr){4-5}
               & \(D_f\) & \(D_r\) & \(D_f\) & \(D_r\) \\
\midrule 
Apple         & 00.00  & 92.25   & 00.00  & 91.12   \\
Baby          & 00.00  & 92.21   & 00.00  & 91.08   \\
Bicycle       & 00.00  & 92.03   & 00.00  & 91.23   \\
Chair         & 00.00  & 92.14   & 00.00  & 91.04   \\
Elephant      & 00.00  & 92.06   & 00.00  & 91.10   \\
\bottomrule
\end{tabular}
}
\caption{Performance of CLIPErase on different patch size CLIP.}
\label{tab:patch14_results}
\end{table}

We conducted additional experiments using the Patch-14 size of CLIP to further evaluate CLIPErase's effectiveness in forgetting specific concepts while preserving other concept in the retain set. The evaluation was performed on multiple forget sets using Zero-Shot Retrieval and Zero-Shot Prediction tasks, measuring performance on both the forget set (\(D_f\)) and the retain set (\(D_r\)).

As shown in Table~\ref{tab:patch14_results}, CLIPErase achieves perfect forgetting across all selected forget classes in both retrieval tasks, demonstrating its capability to fully remove targeted concepts. Meanwhile, the accuracy of the retain set remains stable, averaging 0.9214\% for text retrieval and 0.9111\% for image retrieval. The low variance observed in these results further confirms the robustness of CLIPErase, ensuring reliable and consistent unlearning performance across different patch size CLIP.

These findings highlight that CLIPErase effectively eliminates undesired multimodal associations while maintaining generalization for non-targeted concepts, making it a strong and reliable approach for machine unlearning in multimodal models.

\section{Variance Analysis of Consistency Module}
\label{Variance}

To further analyze the impact of Consistency Module (CM), we conducted variance analysis using the results from different forget set classes shown at Table \ref{tab:stability_variance_results}. The variance of the forget set performance across classes demonstrates the challenges in forgetting certain classes due to their complexity or dependency with other classes. For example, in Zero-Shot Retrieval, the forget set variance is 13.11, compared to the retain set variance of 14.92. In Zero-Shot Prediction, the forget set variance is much lower at 0.038, while the retain set variance is 34.48. These results highlight that CLIPErase performs consistently across various settings, but certain classes like "Mountain," which appear in complex backgrounds or share features with other classes, are harder to forget completely (e.g., 8.19\% accuracy for the forget set in text retrieval). In contrast, more independent classes like "Apple" or "Camel" achieve near-complete forgetting. In summary, RM balances retention and forgetting by prioritizing \(D_r\), while Consistency Module stabilizes optimization and strengthens feature separation, leading to improved performance for both \(D_f\) and \(D_r\). The variance analysis further validates the robustness of CLIPErase in handling diverse class types and highlights the importance of Consistency Module  in achieving consistent performance.

\section{Scalability and Robustness}
\label{Scalabilit}

To demonstrate the scalability and robustness of CLIPErase, we conducted a series of experiments on the Conceptual 12M (CC12M) dataset with varying sizes and complexities of forget sets. These experiments aimed to evaluate CLIPErase's ability to handle single-class, multi-class, and fine-grained forget sets, as well as its performance on larger datasets.

\textbf{Single-Class Forget Set:} First, we evaluated CLIPErase on a single-class forget set to establish baseline performance. We selected "woman" as the forget set target and measured the forgetting performance while maintaining retention for other concepts. The results are shown in Table~\ref{tab:single_class_results}.

\begin{table}[ht]
    \centering
    \footnotesize
    \renewcommand{\arraystretch}{1}
    \scalebox{0.9}{
    \begin{tabular}{l|cc|cc}
    \toprule
    \textbf{Model} & \multicolumn{2}{c|}{\textbf{ZS Prediction (\%)}} & \multicolumn{2}{c}{\textbf{ZS Retrieval (\%)}} \\
    \cmidrule(lr){2-3} \cmidrule(lr){4-5}
                   & \textbf{\(D_f\)} & \textbf{\(D_r\)} & \textbf{\(D_f\)} & \textbf{\(D_r\)} \\
    \midrule
    CLIP & 95.32 & 93.90 & 93.32 & 92.96 \\
    CLIP+GA & 0.00 & 1.44 & 0.11 & 0.02 \\
    CLIP+GradDiff & 90.87 & 93.92 & 91.76 & 93.07 \\
    \rowcolor{lightgray} CLIPErase & 0.33 & 94.19 & 0.42 & 93.19 \\
    \bottomrule
    \end{tabular}
    }
    \caption{Performance comparison across models for Zero-Shot Text and Image Retrieval tasks with a single-class forget set.}
    \label{tab:single_class_results}
\end{table}

CLIPErase significantly outperforms other methods in forgetting the target class while maintaining high retention performance, highlighting its effectiveness in single-class unlearning scenarios.

\textbf{Multi-Class Forget Set:} To further assess scalability on more complex datasets, we conducted additional experiments using a larger and more diverse forget set. Specifically, we selected keywords such as “woman,” “man,” “girl,” “boy,” and “person,” resulting in 42,577 samples in the Forget Set and 77,423 samples in the Retain Set. The unlearning process for this larger dataset was completed within 7 hours on an NVIDIA Tesla V100-SXM2 GPU.

The results of these experiments are summarized in Table~\ref{tab:multi_class_results}. This table compares the performance of the original CLIP model and CLIPErase on both Zero-Shot Retrieval and Zero-Shot Prediction tasks. The comparison demonstrates that CLIPErase effectively forgets the specified set while maintaining high performance on the retain set.

\begin{table}[ht]
    \centering
    \footnotesize
    \renewcommand{\arraystretch}{1}
    \scalebox{1}{
    \begin{tabular}{l|cc|cc}
    \toprule
    \multirow{2}{*}{\textbf{Model}} & \multicolumn{2}{c|}{\textbf{ZS Prediction (\%)}} & \multicolumn{2}{c}{\textbf{ZS Retrieval (\%)}} \\
    \cmidrule(lr){2-3} \cmidrule(lr){4-5}
                                     & \textbf{\(D_f\)} & \textbf{\(D_r\)} & \textbf{\(D_f\)} & \textbf{\(D_r\)} \\
    \midrule 
    \textbf{CLIP}                    & 93.14      & 94.09      & 90.43      & 93.96      \\
    \textbf{CLIPErase}              & 5.75       & 92.67      & 7.08       & 92.38      \\
    \bottomrule
    \end{tabular}
    }
    \caption{Performance of CLIP and CLIPErase on Zero-Shot Text and Image Retrieval tasks with a larger forget set.}
    \label{tab:multi_class_results}
\end{table}

\textbf{Fine-Grained Forget Targets:} Additionally, we conducted experiments to simulate more realistic and fine-grained forgetting scenarios. We set "woman" as the forget set target and evaluated the forgetting performance while maintaining retention for other concepts. The results are shown in Table~\ref{tab:fine_grained_results}.

\begin{table}[h!]
    \centering
    \footnotesize
    \renewcommand{\arraystretch}{1}
    \scalebox{0.9}{
    \begin{tabular}{l|cc|cc}
        \toprule
        \textbf{Model} & \multicolumn{2}{c|}{\textbf{ZS Prediction (\%)}} & \multicolumn{2}{c}{\textbf{ZS Retrieval (\%)}} \\
        \cmidrule(lr){2-3} \cmidrule(lr){4-5}
                       & \textbf{\(D_f\)} & \textbf{\(D_r\)} & \textbf{\(D_f\)} & \textbf{\(D_r\)} \\
        \midrule
        CLIP & 95.32 & 93.90 & 93.32 & 92.96 \\
        CLIP+GA & 0.00& 1.44 & 0.11 & 0.02 \\
        CLIP+GradDiff & 90.87 & 93.92 & 91.76 & 93.07 \\
        \rowcolor{lightgray} CLIPErase & 0.33 & 94.19 & 0.42 & 93.19 \\
        \bottomrule
    \end{tabular}
    }
    \caption{Performance comparison across models for Zero-Shot Text and Image Retrieval tasks with fine-grained forget targets.}
    \label{tab:fine_grained_results}
\end{table}

These experiments demonstrate that CLIPErase maintains computational efficiency and effectively scales to larger and more complex datasets with diverse forget sets. Whether dealing with single-class or multi-class forget sets, CLIPErase consistently achieves robust forgetting while preserving high performance on the retain set. This underscores CLIPErase’s practicality and scalability for real-world applications involving large-scale and multifaceted unlearning tasks.

\section{More Results of Diffusion Models}
\label{diffusion_more}
\begin{figure*}[h!]
\centering
\includegraphics[width=1\textwidth, height=0.5\textheight]{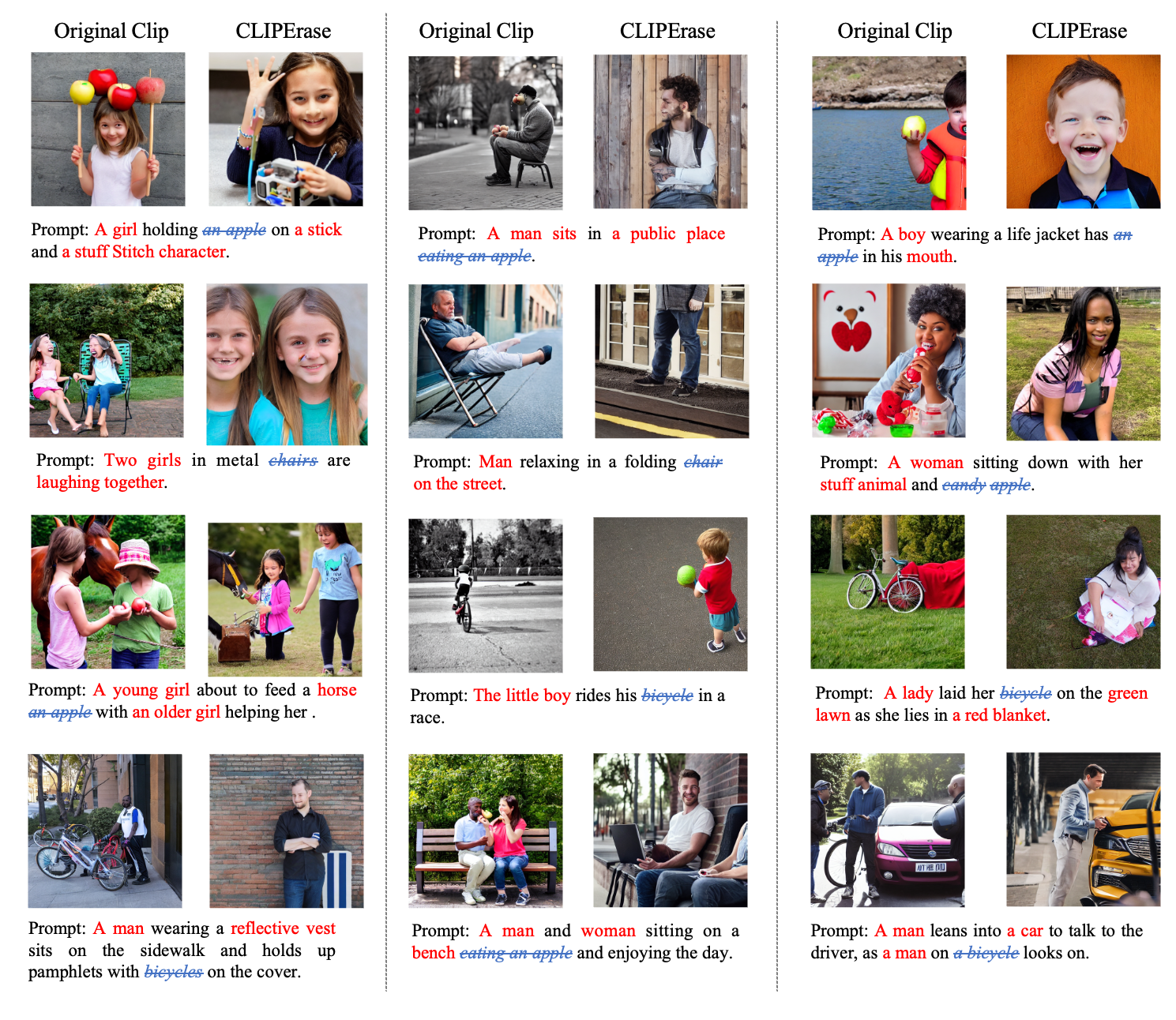}

\caption{Comparison of image generation results using the original CLIP and our CLIPErase model in Stable Diffusion with multi-concept prompts. The prompt represents the input to the diffusion model. Blue text denotes concepts unlearned by CLIPErase, while red text highlights concepts that should be retained.
}

\label{diffmore}
\end{figure*}

As shown in Figure~\ref{diffmore}, we present additional results of images generated using diffusion models with two different encoders. CLIPErase successfully eliminates specific target concepts while retaining other relevant concepts in the generated images.

\section{Discussion and Future Works}
\label{future}

The proposed CLIPErase method has significant potential in practical applications, especially in addressing ethical and legal concerns related to harmful information and biases in multimodal datasets. In tasks like online antisemitism detection, individual components such as text or images may appear harmless on their own, but when combined, they can convey harmful messages. For example, in an image where the text "Even grandma can see what’s going on" seems innocuous at first glance, when paired with an antisemitic image and stereotypical messaging, it transmits damaging, implicit bias. Such hidden biases are especially dangerous in multimodal data. CLIPErase  caneffectively decouple these associations, to forget the harmful links between text and images, thereby mitigating the risk of perpetuating bias.

Additionally, CLIPErase holds great potential in safeguarding user privacy. When users request the deletion of specific personal data, CLIPErase can remove  any associations between their personal information and multimodal content. 

In the future, although our current implementation is focused on the CLIP model, the framework can be extended to any modality or multimodal pretrained model, not just CLIP. This broader applicability would enable flexible unlearning across a wide range of systems, making the approach more versatile. Additionally, we aim to apply CLIPErase to Generative AI systems, such as Multimodal Large Language Models (MLLMs), 
where CLIP-based encoders are widely used. By unlearning at the encoder level, CLIPErase can help address the growing challenges in Generative AI, including the generation of private, malicious, or illegal content, the continuation of biases, and even the risk of weaponizing these models. Our approach can serve as a safeguard, correcting problematic associations and enhancing user privacy, thus providing a safer and more ethical experience in the future of AI development.

\end{document}